\documentclass[final]{tmi99}
\usepackage{gb4e}

\title{A Unified Example-Based and Lexicalist Approach to Machine Translation}
\author{Davide Turcato \ \ Paul McFetridge \ \ Fred Popowich and Janine Toole}

\institute{Natural Language Laboratory, School of Computing Science,
Simon Fraser University\\8888 University Drive, Burnaby, British
Columbia, V5A 1S6, Canada}

\address{and\\Gavagai Technology\\P.O. 374, 3495 Cambie Street, Vancouver,
British Columbia, V5Z 4R3, Canada}

\email{\{turk,mcfet,popowich,toole\}@cs.sfu.ca}

\begin{document}
\maketitle

\begin{abstract}
We propose an approach to Machine Translation that combines the ideas
and methodologies of the Example-Based and Lexicalist theoretical
frameworks. The approach has been implemented in a multilingual
Machine Translation system.
\end{abstract}

\section{Introduction}

Human translation is a complex intellectual activity and accordingly
Machine Translation (henceforth MT) is a complex scientific task,
involving virtually every aspect of Natural Language Processing. Many
approaches have been proposed, each of them inspired by some insight
about translation. Each approach has its own merit, accounting for
some aspect of translation better than other approaches, but typically
each approach's advantages are countered by weaknesses in other
respects. The real challenge is combining different approaches and
insights into a comprehensive whole. To this end it is important to
compare different approaches, for two reasons:

\begin{enumerate}

\item It is important to see to what extent differences are
substantial or notational. Sometimes different approaches look at the
same subject from different viewpoints, or use different
representations, but a formal analysis shows that they are
equivalent. This was the case with many formal systems (categorial and
phrase structure grammars, finite state machines and regular grammars,
explanation-based generalization and partial evaluation, etc.). In
other cases differences have been demonstrated to be matters of degree
(for instance, in the field of MT, transfer and interlingua
approaches).

\item It is important to see to what extent different approaches are
mutually exclusive, or whether they can be integrated into one system
that encompasses all of them.

\end{enumerate}

In this paper we examine the Example-Based and the Lexicalist
approaches to MT. Despite the differences between their paradigms and
methods, we argue that:

\begin{enumerate}

\item the kind of linguistic resources the two approaches use largely
overlap and can be expressed in the same notation;

\item a unified MT architecture can be proposed that encompasses the
methodologies of both approaches.

\end{enumerate}

\section{The Two Approaches}

\subsection{Example-Based MT}

\namecite[198]{Arnold:94MTAIG} outline Example-Based MT (henceforth
EBMT) as follows: ``The basic idea is to collect a bilingual corpus of
translation pairs and then use a best match algorithm to find the
closest example to the source phrase in question. This gives a
translation template, which can then be filled in by word-for-word
translation.''

The paradigm of \emph{translation by analogy} was first introduced by
\namecite{Nagao:84AHI}. In that paper Nagao advocates the use of raw,
unanalyzed bilingual data, claiming that linguistic data are more
durable than linguistic theories, thus constituting a steadier ground
for MT systems. He proposes the use of an unannotated database of
examples (possibly collected from a bilingual dictionary) and a set of
lexical equivalences simply expressed in terms of word pairs (except
for verb equivalences, which are expressed in terms of case
frames). The matching process is mainly focused on checking the
semantic similarity between the lexical items in the input sentence
and the corresponding items in the candidate example.

Many variations and extensions to Nagao's ideas followed, under
different names and acronyms: Example-Based Machine Translation (EBMT,
\citeboth{Sumita:ACL91}), Memory-Based Translation (MBT,
\citeboth{Sato:COLING90}), Transfer-Driven Machine Translation (TDMT,
\citeboth{Furuse:COLING92}), Case-Based Machine Translation (CBMT,
\citeboth{Kitano:IJCAI93a}), etc. There is not always agreement about
the usage of such names: for instance, some authors use EBMT and MBT
interchangeably, while others keep them distinct. However, all these
approaches share the basic idea described above. The main directions
in which Nagao's original model has been extended are the following:

\begin{enumerate}

\item Augment the example database with linguistic annotations and,
accordingly, perform some linguistic analysis on the input before the
matching phase. \namecite{Sato:COLING90} store examples in the form of
pairs of word-dependency trees, along with a set of `correspondence
links'. For instance, the English-Japanese pair of sentences in
(\ref{eat-sentence}) is represented as the Prolog facts in
(\ref{eat-tree}):

\begin{exe}

\ex \label{eat-sentence}

\begin{xlist}

\ex \citeform{He eats vegetables.}

\ex \citeform{Kare ha yasai wo taberu.}

\end{xlist}

\end{exe}

\begin{exe}

\ex \label{eat-tree}

\begin{verbatim}
ewd_e([e1,[eat,v],
          [e2,[he,pron]],
          [e3,[vegetable,n]]]).

jwd_e([j1,[taberu,v],
          [j2,[ha,p],
              [j3,[kare,pron]]],
          [j4,[wo,p],
              [j5,[yasai,n]]]]).

clinks([[e1,j1],[e2,j3],[e3,j5]]).
\end{verbatim}

\end{exe}

\namecite{Kitano:IJCAI93a} proposes the annotation of examples with
morphological information for words and then suggests splitting the
source and target sentences into \emph{segments}, i.e. continuous
sequences of words. He then proposes to annotate examples with a
\emph{segment map}, i.e. a correspondence between segments in the
source and target sentences, in a similar fashion to what
\namecite{Sato:COLING90} do with word-dependency sub-trees.

\item Explicitly store templates in the bilingual database, instead of
sentences \cite{Kaji:COLING92}. A template is a sentence where some
phrases have been replaced by variables, annotated with linguistic
information. For instance:

\begin{exe}

\ex \label{template} \texttt{X[PRON] eats Y[NP] <-> X[PRON] ha Y[NP] wo taberu}

\end{exe}

Templates are learnt from pairs of sentences by parsing them, in order
to perform correct replacement of words or phrases with annotated
variables, and to obtain cross-linguistic variable-sharing.

\namecite{Furuse:COLING92} propose encoding different kinds of
bilingual correspondences in the database: string-level
correspondences (i.e. plain phrase pairs), pattern-level
correspondences (pairs of templates containing variables),
grammar-level correspondences (pairs of templates containing variable
annotated with syntactic categories, like those proposed by
\citeauthor{Kaji:COLING92}). Moreover, they associate a source
expression with several target expressions, each of which is provided
with a set of examples that show the contexts in which each target
expression can be correctly used. For instance:

\begin{exe}

\ex \label{tdmt-example}\begin{verbatim}
X wo o-negaishimasu <->
	may I speak to X'  ((jimukyoku{office}), ...)
	please give me X'  ((bangou{number}), ...)
\end{verbatim}

\end{exe}

where \texttt{X'} is the translation of \texttt{X} and each
\texttt{(X\{X'\})} pair in parentheses is an instantiation of such a
translation pair.

\item Obtain the translation of a complete sentence by utilizing more
than one translation example and combine some fragments of them. For
instance, \namecite{Sato:COLING90} show how to obtain the translation
(\ref{tiling1}), given the examples (\ref{tiling2}) and
(\ref{tiling3}).

\begin{exe}

\ex \label{tiling1} \citeform{He buys a book on international politics
$\leftrightarrow$\\ Kare ha kokusaiseiji nitsuite kakareta hon wo kau}

\ex \label{tiling2} \citeform{\textbf{He buys} a notebook
$\leftrightarrow$\\ \textbf{Kare ha} nouto \textbf{wo kau}}

\ex \label{tiling3} \citeform{I read \textbf{a book on international
politics} $\leftrightarrow$\\ Watashi ha \textbf{kokusaiseiji nitsuite
kakareta hon} wo yomu}

\end{exe}

For an input sentence, they construct a \emph{matching expression},
i.e. a pointer to a \emph{translation unit}. A translation unit is a
word-dependency sub-tree to be found in the example database (the
\texttt{e$_{1}$, \ldots, e$_{n}$, j$_{1}$, \ldots, j$_{m}$} of example
(\ref{eat-tree})). Such pointers can be optionally followed by a list
of commands for deletion/replacement/adjunctions of nodes dominated by
the node pointed to. The replaced or adjoined elements are other
matching expressions. For instance, given (\ref{eat-tree}), a matching
expression for the sentence (\ref{me-sentence}) might be
(\ref{me-me}).

\begin{exe}

\ex \label{me-sentence} \citeform{He eats mashed potatoes}

\ex \label{me-me} \texttt{[e1,[r,e3,[e$_{x}$]]]}

\end{exe}

\noindent which represents the tree obtained by replacing (\texttt{r}
for `replace') \texttt{e3} with \texttt{e$_{x}$} in \texttt{e1}. In
turn, \texttt{e$_{x}$} is a pointer to a sub-tree for \citeform{mashed
potatoes} in some other example.

As several matching expressions can be candidates for the same input
sentence, \citeauthor{Sato:COLING90} define a scoring system for
competing translation units, based on their length (the longer, the
better) and the semantic similarity between their contexts, i.e. the
input sentence and the example from which the translation unit is
taken (the more similar, the better).

\end{enumerate}

\subsection{Lexicalist MT}

Lexicalist MT (henceforth LMT) is a variant of the transfer approach
to MT. In LMT transfer is a mapping between bags of lexical items,
instead of trees \cite{Whitelock:94SB}.

The first step of the translation process is the analysis of an input
sentence. Such analysis is performed on a purely monolingual basis,
independently from considerations of translation direction and
language pair. The same kind of declarative grammars are used for
parsing and generation. Moreover, grammars tend to follow a
lexicalist, sign-based\footnote{Following the Saussurean approach
taken in HPSG, we define signs as ``structured complexes of
phonological, syntactic, semantic, discourse and phrase-structural
information'' \cite[15]{Pollard:94HPSG}.} approach
\cite{Pollard:94HPSG}. Grammar rules are reduced to a small number of
general rule schemata. Lexical items are multidimensional signs
containing all the information about their modes of combination
(subcategorization, head-modifier relations, etc.). As a result of
parsing, lexical items are instantiated with indices expressing their
interdependencies with other lexical items in the sentence.

Transfer is a mapping from a bag of instantiated source lexical items
resulting from parsing to a corresponding bag of target lexical
items. Bilingual knowledge is reduced to a bilingual lexicon,
augmented with cross-linguistic correspondences in the form of equated
variables. Transfer is performed by finding a set of bilingual entries
that covers the source bag. The target bag is comprised of the target
sides of the selected bilingual entries. Phrasal and idiomatic
expressions are accounted for by multi-word bilingual entries, where
lexical items on either side have no inherent order and can be
discontinuous in the input or output sentence. A schematic bilingual
entry is shown in (\ref{eat-bilex}), where subscripts represent
indices encoding word dependencies.

\begin{exe}

\ex \label{eat-bilex} \texttt{\emph{eat}:v$_{a,b,c}$ $\leftrightarrow$
\emph{taberu}:v$_{a,d,e}$ \& \emph{ha}:p$_{d,b}$ \&
\emph{wo}:p$_{e,c}$}

\end{exe}

Generation orders the target lexical items into a grammatical
sentence, according to a target grammar and to the constraints
expressed by the indices instantiated on target lexical items as a
result of transfer.

\subsection{Comparison}

It is interesting to note that the introduction of both EBMT and LMT
was motivated by the rejection of structural transfer, due to its
inadequacy to cope with structurally divergent languages like English
and Japanese (\citeboth[179]{Nagao:84AHI};
\citeboth[343--345]{Whitelock:94SB}). However, the two approaches
differ in the way they avoid the recursive traversal of an analysis
tree structure in transfer.

In EBMT structural transfer is avoided by adopting the following
guidelines:

\begin{enumerate}

\item Sentence-level correspondences are covered via an explicit
stipulation of all such possible correspondences in the bilingual
knowledge base. Therefore EBMT advocates a bilingual knowledge base
stating equivalences between the maximal translation units,
i.e. sentences (or sentence templates).

\item Given such flat structure of the bilingual database, no deep
linguistic analysis is required. Transfer is performed by looking up
the bilingual database for a suitable match to the input
sentence. Linguistic analysis is only performed to the extent it is
necessary to effectively perform the template matching.

\end{enumerate}

In LMT the following guidelines are adopted:

\begin{enumerate}

\item A full linguistic analysis is performed on input sentences. As a
result of the lexicalist, sign-based approach to parsing and the use
of indices to represent dependencies among lexical items, individual
lexical items contain all the information about their structural
relationships with the other lexical items.

\item Given that all the information about a sentence structure is
stored in lexical items, transfer can be reduced to a mapping of a bag
of source lexical items onto a bag of target lexical items.  Therefore
LMT advocates a bilingual knowledge base stating equivalences between
the minimal translation units, i.e. lexical items. Information about
word order is also dropped from transfer, as it is considered a
monolingual issue, accounted for by the linear precedence constraints
expressed in grammar rules. The dependencies expressed by indices are
the only information that must be necessarily transferred from source
to target lexical items.

\end{enumerate}

\section{A Unified Bilingual Knowledge Base}

Despite the different and somehow antithetic architectures, the kind
of bilingual resources required by the two approaches tend to
converge. We show that it is possible to define a bilingual knowledge
base in such a way that it can serve both approaches.

At a formal level, it can be shown that the kind of information used
by the two approaches largely overlaps. The information needed for
EBMT systems can be adequately expressed in a LMT notation. We list
here some parallelisms:

\begin{enumerate}

\item Case frames as used in EBMT correspond to subcategorization
frames in LMT. Therefore an EBMT case frame is equivalent to an LMT
verb lexical entry. Aside from word order issues, which will be dealt
with later, the same holds for templates where some arguments are left
unspecified, as a comparison between (\ref{template}) and
(\ref{eat-bilex}) shows.

\item A comparison between (\ref{eat-tree}) and (\ref{eat-bilex})
shows that \quotecite{Sato:COLING90} word-dependency trees contain the
same information as LMT bilingual entries: words, grammatical
descriptions, monolingual dependencies, cross-linguistic
correspondences.

\item TDMT templates correspond to sets of bilingual entries.

\end{enumerate}

A bilingual lexicon as used in LMT can adequately represent all the
information needed in EBMT. Therefore, we advocate the use of the LMT
notation as a theory neutral knowledge representation language that
can equally support EBMT and LMT bilingual knowledge bases. The
adoption of such notation does not commit one to using one or the
other approach. Such a bilingual resource is close in spirit to the
kind of Bilingual Knowledge Bank advocated by
\namecite{Sadler:COLING90}.

The neutrality of the proposed notation also relies on the fact the
notation's semantics is underspecified. The notation is such that it
can be interpreted in different ways, in developing and using a
knowledge base. Particularly, a bilingual entry's source or target
side can be interpreted as either a bag or a sequence. In the latter
case, word order is relevant in matching some input with a bilingual
entry. A further constraint on the matching procedure may be the
requirement that input words matching bilingual entry items must be
contiguous in the sentence. If both order and contiguity constraints
are activated, then a bilingual knowledge base is interpreted as a
knowledge base of sentences (or phrases), as in
\quotecite{Nagao:84AHI} original proposal, or segments, as proposed by
\namecite{Kitano:IJCAI93a}. If the order constraint is activated and
the contiguity constraint is dropped, then bilingual entries represent
templates. If both constraints are dropped, then bilingual entries
represent word-dependency trees or LMT lexical bags. Therefore the
same notation can be used with different ideas in mind and, to some
extent, the same knowledge base can be reused under different
interpretations.

Our experience in large scale bilingual lexical development
(English-Spanish) showed that the commitment to a specific semantics
may even be changed after a bilingual knowledge base has been
developed, if some conventions in writing entries are observed. We
remarked that our lexicographers spontaneously used the obvious
convention of writing bilingual entries items in the same order in
which words appear in sentences. With some exceptions (for instance,
Spanish verbs accompanied by clitic pronouns), the order in which
bilingual entry items can appear in sentences turned out to be unique
in most cases. This gave us the choice of interpreting our bilingual
entries as either bags or sequences, which was an option unforeseen at
the beginning. It is also possible to choose a mixed semantics,
e.g. interpreting source sides as sequences and target sides as
bags. Different considerations come into play for different languages
and translation directions. For instance, the order constraint might
be appropriate for a language with a relatively fixed word order, but
not for one with a relatively free word order (even more so, when the
language at hand is used as a source language).

The notation can also be extended to contain explicit place-holders
for missing elements, thus resembling templates more closely. For an
illustration of such an extension see \cite{Turcato:RANLP97}.

Besides notation, a second issue is the actual information that the
two approaches require of a bilingual knowledge base. As noted above,
EBMT tends to require equivalences between maximal translation units,
i.e. sentences, while LMT tends to require equivalences between minimal
units, i.e. lexical items. Such a divergence is actually less dramatic
if we take a closer look at the issue. To this end we introduce a
distinction between two different purposes of an example database.

\begin{enumerate}

\item Examples provide information about sentence well-formedness. The
required amount of such information is inversely proportional to the
amount of linguistic analysis performed on the input. At one extreme,
we have the case that no analysis is performed. In this case, all the
possible sentences should be listed in the bilingual database. The
next level is when input words are assigned syntactic categories. In
this case, sentences in the bilingual database can be either replaced
by or used as templates. At the opposite extreme is the case where a
complete linguistic analysis is performed. In this case, examples are
no longer needed to provide information about well-formedness. Lexical
equivalences are sufficient. For instance, if we assume that a
sentence input is analyzed into a word-dependency tree, then a
bilingual example like (\ref{eat-tree}) can be replaced by a set of
bilingual lexical entries without any loss of information (provided,
as is assumed by a lexicalist approach, that each lexical item
contains information about the arguments it subcategorizes
for). Therefore a lexicalist approach can be seen as the lower bound
on a continuous scale of different EBMT approaches, depending on the
amount of linguistic analysis performed.

\item Examples provide information about non-compositional
translations (e.g. idioms) and contrastive information about different
ways in which a word is translated in different contexts
(sense-ambiguous words). This is the case of examples like
(\ref{tdmt-example}), for instance. This kind of information is
equally required by EBMT and LMT systems, regardless of the chosen
approach to linguistic analysis, and needs to be expressed in either
case by multi-word bilingual entries.

\end{enumerate}

To sum up, the required information is the same in EBMT and LMT, to
some extent. The extent of the residual difference is a matter of
degree of linguistic analysis performed by the system.

\section{A Unified Architecture}

\namecite[201]{Arnold:94MTAIG} suggest that ``there is no radical
incompatibility between example-based and rule-based approaches, so
that the challenge lies in finding the best combination of techniques
from each. Here one obvious possibility is to use traditional
rule-based transfer as a fall back, to be used only if there is no
complete example-based translation.'' Rather than proposing a
multi-engine approach with a duplication of resources, we propose a
single architecture that encompasses the two approaches and integrates
the basic tenets of both.

A common characteristic of all EBMT approaches is that the translation
process is driven by the content of the bilingual knowledge base. The
core operation of all such approaches is the match of an input
sentence against examples. It is the result of such a match that
drives further computation, in terms of calculating similarity,
replacing items in the chosen example, combining fragments of
different examples (this prioritization of transfer is made explicit
in approaches like TDMT). On the contrary, in LMT different
translation steps are clearly separated. As pointed out, parsing is
performed on a purely monolingual ground, regardless of the specific
translation flow in which it occurs. We propose a translation
architecture that combines the advantages of the two approaches, by
using bilingual information to drive the translation process, while
preserving the modularity of the system.

A bilingual knowledge base as described above is not only a source of
bilingual information, but it also encodes a considerable amount of
monolingual linguistic information, on either side. A multi-word
bilingual entry gives syntactic and semantic information about the
analysis of phrasal expressions, collocations and idioms. Even
single-word entries give clues about the analysis of lexically
ambiguous items.

We propose to take advantage of this source of information to drive
the parsing process of an input sentence. Given the search space
defined by the monolingual lexicon and grammar, the information
contained in the bilingual lexicon is used to prioritize certain
analyses over others\footnote{A similar idea has been proposed by
\namecite[80]{Kinoshita:SBIA98} in a different theoretical
framework.}. A bilingual lexicon lookup before parsing offers a
partial analysis (in terms of lexical disambiguation, dependencies
and, optionally, word order), which is tried before any other
hypothesis supported by the monolingual lexicon and grammar. If, for
instance, chart parsing is in use, the example-based approach amounts
to prioritizing edges in the chart agenda. Moreover, edges licensed by
the same multi-word bilingual entry are assigned a common identifier,
so as to ensure that they all fail or succeed together.

When several bilingual entries apply, they are prioritized by the
cardinality of their side being used. This sorting mechanism
implements a kind of `elsewhere condition': more specific entries
override more general ones. This device can be regarded as a
lexicalist implementation of the scoring mechanism described by
\namecite{Sato:COLING90}, according to which longer translation units
are preferred over shorter ones.

We show how the translation process works with an English-Spanish
example:

\begin{exe}

\ex \label{eng-sent} \emph{They cut back on investments}

\end{exe}

Let's assume that the English lexicon and the bilingual lexicon
contain, respectively, the following (simplified) entries:

\begin{center}
\begin{tabular}{l}
\texttt{\emph{back}:adv$_{a}$}		\\
\texttt{\emph{back}:n$_{a}$}		\\
\texttt{\emph{cut}:iv$_{a,b}$}		\\
\texttt{\emph{cut}:iv$_{a,b,c}$}	\\
\texttt{\emph{cut}:tv$_{a,b,c}$}	\\
\texttt{\emph{investment}:n$_{a}$}	\\
\texttt{\emph{on}:p$_{a,b}$}		\\
\end{tabular}
\end{center}

\begin{center}
\begin{tabular}{rll}
a. & \texttt{\emph{back}:adv$_{a}$}							& \texttt{$\leftrightarrow$\emph{atr\'{a}s}:adv$_{a}$}					\\
b. & \texttt{\emph{back}:n$_{a}$}							& \texttt{$\leftrightarrow$\emph{espalda}:n$_{a}$}					\\
c. & \texttt{\emph{cut}:iv$_{a,b}$}							& \texttt{$\leftrightarrow$\emph{cortar}:iv$_{a,b}$}					\\
d. & \texttt{\emph{cut}:iv$_{a,b,c}$ \& \emph{back}:adv$_{c}$}				& \texttt{$\leftrightarrow$\emph{hacer}:tv$_{a,b,d}$ \& \emph{econom\'{\i}a}:n$_{d}$}	\\
e. & \texttt{\emph{cut}:iv$_{a,b,c}$ \& \emph{back}:adv$_{c}$ \& \emph{on}:p$_{a,d}$}& \texttt{$\leftrightarrow$\emph{reducir}:tv$_{a,b,d}$}					\\
f. & \texttt{\emph{cut}:tv$_{a,b,c}$}							& \texttt{$\leftrightarrow$\emph{cortar}:tv$_{a,b,c}$}					\\
g. & \texttt{\emph{investment}:n$_{a}$}							& \texttt{$\leftrightarrow$\emph{inversi\'{o}n}:n$_{a}$}				\\
h. & \texttt{\emph{on}:p$_{a,b}$}							& \texttt{$\leftrightarrow$\emph{en}:p$_{a,b}$}
\end{tabular}
\end{center}

Given such monolingual and bilingual lexical entries, we show below
all the possible ways in which the input sentence (\ref{eng-sent}) can
be grammatically covered in parsing and correspondingly translated (we
omit details about \emph{they}, which is syntactically unambiguous and
is dropped in Spanish). The solutions are listed in no specific
order. Note that more than one translation can be given for the same
parse, depending on what bilingual entries are used. Conversely,
different parses can result in the same translation. At the end of
each line we also indicate what bilingual entries have been
used.

\begin{center}
\begin{tabular}{ccccll}
\emph{cut}		& \emph{back}		& \emph{on}		& \emph{investments}	&								&		\\
\texttt{iv$_{a,b}$}	& \texttt{adv$_{a}$}	& \texttt{p$_{a,c}$}	& \texttt{n$_{c}$}	& $\rightarrow$	\emph{cortan atr\'{a}s en las inversiones}	& \{cahg\}	\\
\texttt{tv$_{a,b,c}$}	& \texttt{n$_{c}$}	& \texttt{p$_{a,d}$}	& \texttt{n$_{d}$}	& $\rightarrow$	\emph{cortan espalda en las inversiones}	& \{fbhg\}	\\
\texttt{tv$_{a,b,c}$}	& \texttt{n$_{c}$}	& \texttt{p$_{c,d}$}	& \texttt{n$_{d}$}	& $\rightarrow$	\emph{cortan espalda en las inversiones}	& \{fbhg\}	\\
\texttt{iv$_{a,b,c}$}	& \texttt{adv$_{c}$}	& \texttt{p$_{a,d}$}	& \texttt{n$_{d}$}	& $\rightarrow$	\emph{hacen econom\'{\i}as en las inversiones}	& \{dhg\}	\\
\texttt{iv$_{a,b,c}$}	& \texttt{adv$_{c}$}	& \texttt{p$_{a,d}$}	& \texttt{n$_{d}$}	& $\rightarrow$	\emph{reducen las inversiones}			& \{eg\}	\\
\end{tabular}
\end{center}

Note that in our specific example it is irrelevant whether the order
and contiguity constraints are enforced on the bilingual lexicon. The
parsing strategy we propose first tries to find a parse consistent
with the source side of bilingual entry (e), the longest
available. Therefore, assuming that no failure occurs, the first
translation returned is \emph{reducen las inversiones}, which is the
most correct. If a failure occurs anywhere down the path for all the
parses covered by the source side of bilingual entry (e), the source
side of bilingual entry (d) is tried next. Therefore, \emph{hacen
econom\'{\i}as en las inversiones} would be the second translation
returned. When the bilingual lexicon offers no way of prioritizing
among parsing hypotheses, any other available prioritization mechanism
can still be used. Also, in using a bilingual lexicon, different
strategies could be used. For instance, an alternative to using the
longest match first, would be to look for the cover that uses the
fewest number of bilingual entries.

As discussed above, \namecite{Sato:COLING90} also use a second scoring
mechanism, based on similarity between the input sentence and the
candidate translation units. A lexicalist counterpart of such a
mechanism would amount to a word sense disambiguation module, provided
that senses are associated with words in the bilingual lexicon. In
fact, the problem of choosing the right translation for a word or
phrase that can be translated in different ways amounts to choosing
the correct word sense for that word or phrase. This, in turn, is
customarily done in the word sense disambiguation literature by
looking at the context in which the word or phrase occurs
(e.g. \citeboth{Resnik:WVLC3-95}; \citeboth{Yarowsky:ACL95}), thus
paralleling \quotecite{Sato:COLING90} idea. Although we have not
implemented any such word sense disambiguation module, it would be
straightforward to incorporate such a module in the system
architecture without affecting the other modules. As for associating
senses with lexical items in the bilingual lexicon, a method for
automatically selecting and ranking bilingual entries (unannotated for
sense), based on an input word's sense and context, has been proposed
by \namecite{Sanfilippo:TMI97}.

Besides the monolingual considerations discussed above, this approach
also has the advantage of biasing parsing towards analyses that are
supported by the bilingual lexicon. An analysis, even a correct one,
is useless if the transfer component does not have the means to map it
onto a target representation. Therefore it is a practical choice to
prioritize those analyses that are amenable to a successful transfer.

The proposed approach, which has been incorporated into a large scale
MT system \cite{Popowich:TMI97}, does not affect the results provided
by the system, it only affects the order in which they are
provided. Of course, if a system returns the first solution found, the
system behavior indeed changes.

\section{Conclusion}

A knowledge representation format and a system architecture have been
proposed that allow an effective integration of Example-Based and
Lexicalist approaches to MT into a unified approach, which we call
Example-Based Lexicalist Machine Translation (EBLMT). This approach
combines the advantages of each approach. From the point of view of
LMT, it uses bilingual knowledge to drive parsing, providing
additional information to solve syntactic ambiguities and prioritizing
the parsing agenda in a more efficient way. From the point of view of
an EBMT system like \cite{Sato:COLING90}, for instance, it allows the
removal of the bilingual database's redundancy coming from the overlap
of examples. Moreover, the flexibility of the knowledge representation
format and the modularity of the architecture allow a system to work
in different modalities, by simply setting some system parameters.

\section*{Acknowledgements}

This research was supported by a Collaborative Research and
Development Grant from the Natural Sciences and Engineering Research
Council of Canada (NSERC), and by the BC Advanced Systems Institute
(ASI).

\bibliographystyle{tmi}

% \bibliography{99tmi-eblmt}

\end{document}